%% file: aaai25.tex
\documentclass[letterpaper]{article} 
\usepackage{aaai25}  
\usepackage{times}  
\usepackage{helvet}  
\usepackage{courier}  
\usepackage[hyphens]{url}  
\usepackage{graphicx} 
\usepackage{natbib}  
\usepackage{caption} 
\usepackage{algorithm}
\usepackage{algorithmic}

\usepackage{newfloat}
\usepackage{listings}
\DeclareCaptionStyle{ruled}{labelfont=normalfont,labelsep=colon,strut=off} 
\floatstyle{ruled}
\newfloat{listing}{tb}{lst}{}
\floatname{listing}{Listing}
\title{Beyond Accuracy: On the Effects of Fine-tuning \\ Towards Vision-Language Model's Prediction Rationality}
\author{
    Qitong Wang,
    Tang Li,
    Kien X. Nguyen,
    Xi Peng
}
\affiliations{

    DeepREAL Lab, Department of Computer \& Information Sciences, University of Delaware \\
    \{wqtwjt, tangli, kxnguyen, xipeng\}@udel.edu
}

\usepackage{bibentry}

\usepackage{multirow}
\usepackage{parskip}
\usepackage{booktabs}
\usepackage{amssymb}
\usepackage{amsmath}
\usepackage{xcolor}

\begin{document}

\maketitle

\input{sec_ours/0_abstract}

\input{sec_ours/1_introduction}
\input{sec_ours/2_3_4.1_experimental_setup}
\input{sec_ours/4.2_weakness}
\input{sec_ours/4.3_strength}
\input{sec_ours/4.4_ood}
\input{sec_ours/4.5_ablation_study}
\input{sec_ours/5_related_work}
\input{sec_ours/6_conclusion}

\bibliography{aaai25}

\input{sec_ours/suppl}

\end{document}

%% file: sec_ours/0_abstract.tex
\begin{abstract}
\normalsize
Vision-Language Models (VLMs), such as CLIP, have already seen widespread applications.
Researchers actively engage in further fine-tuning VLMs in safety-critical domains. 
In these domains, prediction rationality is crucial: \textit{the prediction should be correct and based on valid evidence}.
Yet, for VLMs, the impact of fine-tuning on prediction rationality is seldomly investigated. 
To study this problem, we proposed two new metrics called \textit{Prediction Trustworthiness} and \textit{Inference Reliability}.
We conducted extensive experiments on various settings and observed some interesting phenomena.
On the one hand, we found that the well-adopted fine-tuning methods led to more correct predictions based on invalid evidence.
This potentially undermines the trustworthiness of correct predictions from fine-tuned VLMs.
On the other hand, having identified valid evidence of target objects, fine-tuned VLMs were more likely to make correct predictions.
Moreover, the findings are also consistent under distributional shifts and across various experimental settings.
We hope our research offer fresh insights to VLM fine-tuning.
\end{abstract}
\begin{links}
    \link{Code}{https://github.com/deep-real/vlm-pred-rationality}
\end{links}

%% file: sec_ours/1_introduction.tex
\section{Introduction}
\label{sec:intro}

Vision-Language Models (VLMs), such as CLIP~\cite{clip}, have recently begun to see widespread adoption in high-stakes applications, such as healthcare~\cite{medclip} and autonomous driving~\cite{clip2scene}.
A common practice in utilizing VLMs involves undertaking further fine-tuning~\cite{flyp, wise_ft, medclip} in these models to their specific tasks rather than training deep models from scratch.
While existing studies have evaluated mainstream fine-tuning methods, they have primarily focused on prediction accuracy~\cite{kumar2022calibrated, wortsman2022robust, flyp}, overlooking an essential aspect: \textit{prediction rationality}, where model predictions should not only be accurate but also grounded in valid evidence.
Besides, the current academic community widely accepts that ``clearly explaining a rationale for a classification decision to an end-user can be as important as the decision itself.''~\cite{hendricks2016generating}
One significant reason is that neglecting the model's prediction rationality will cause severe consequences in safety-critical domains.
For example, doctors employ a fine-tuned VLM which can accurately predict the presence of cancer tumors from X-ray images, to help decision-making.
If its predictions are based on erroneous reasons: the input's background instead of tumor region, doctors will lack trust in fine-tuned VLM, leading them to disregard the usage of the model. 
Therefore, in this paper, we study a crucial yet seldom investigated question: \textit{how do mainstream fine-tuning methods affect the rationality of VLM predictions?}

To systematically study this question, we propose two new metrics to evaluate the rationality of VLM predictions after fine-tuning: 
(1) Prediction Trustworthiness (PT): the ratio of correct predictions with valid evidence overall correct predictions. 
(2) Inference Reliability (IR): the percentage of correct predictions given that the model has identified valid evidence of target objects. 
To assess whether the model focuses on valid evidence for the image classification task, we measure if the generated explanation heatmap from VLMs focuses on the target objects, based on the ``Relevant Mass Accuracy (RMA)'' score~\cite{es_eval2}.
We study the mainstream methods including ``Zero-Shot'' (ZS), ``Linear-Probing'' (LP), ``Finetune Like CLIP Pretrain'' (FLCP), and standard ``Fine-tuning'' (FT). 
We conducted extensive experiments and have obtained novel and consistent findings.
Our results reveal that widely used fine-tuning methods exhibit significant limitations, yet they also possess certain advantages.
Our key findings are summarized as follows:

\textbf{Will mainstream fine-tuning methods hurt the rationality of VLM predictions? Surprisingly yes!}
With our proposed ``Prediction Trustworthiness'' metric, fine-tuning results in more appearance of samples with correct predictions based on invalid evidence than zero-shot, making the correct predictions untrustworthy. 
For instance, with the ALBEF-ViT-B/16 model, compared with ZS, the PT scores of LP, FLCP, and FT drop $17.2\%$, $13.85\%$ and $27.31\%$ respectively, on CalTech-101~\cite{caltect101} dataset, despite improving prediction accuracies.
And with the CLIP-ViT-B/16 model, compared with ZS, the PT scores of LP, FLCP, and FT drop $6.4\%$, $5.65\%$, and $4.07\%$ respectively, on ImageNet-1K~\cite{imagenet} dataset.
Notably, existing work~\cite{flyp} highlights the effectiveness of fine-tuning for VLMs, asserting that FLCP consistently improves prediction accuracy and should be considered the ``standard'' method for fine-tuning CLIP. 
However, our findings suggest that this conclusion does not hold when evaluating the rationality of VLM predictions.
This discrepancy underscores the importance of considering different possibilities when evaluating prediction rationality.

\textbf{Will valid evidence help enhance predictions made by fine-tuned VLMs? Yes.}
Using our ``Inference Reliability'' metric, we find that when VLMs focus on valid evidence of target objects, the prediction accuracy of fine-tuned VLMs improves.
For example, in the ImageNet-1K dataset, with the CLIP-ViT-B/16 model, LP, FLCP, and FT outperform ZS in IR scores by $12.6\%$, $8.67\%$, and $16.92\%$ respectively.
Existing works~\cite{kumar2022calibrated, wortsman2022robust, flyp}, which study the positive impacts of VLM fine-tuning, are limited to the prediction accuracies. 
Our research provides insights into the impact of fine-tuning VLMs from a novel perspective, highlighting the benefits of fine-tuning in terms of enhancing prediction rationality.

\textbf{Will out-of-distribution data change our observations? No.}
There is a critical need to make sure that models work reliably in real-world situations, where the data distribution they encounter might be different from what they were trained on.
For instance, the model must maintain stability and effectiveness in autonomous driving applications across various weather conditions.
In parallel, previous work~\cite{clip} has demonstrated the remarkable predictive performance of CLIP in both in-distribution and out-of-distribution data. 
Therefore, we discuss how our observations might change in the context of out-of-distribution data.
We find that all our findings remain consistent across various types and magnitudes of distributional shifts, as demonstrated through experiments in ImageNet-C~\cite{imagenet-c}.

Lastly, we conducted ablation studies to verify the consistency of our findings, which remain consistent across various experimental settings, including different training optimizers, learning rates, explanation heatmap methods, and fine-tuning techniques such as prompt tuning~\cite{coop} and adapter tuning~\cite{tipa}.

Our contribution lies in discovering new findings through extensive experiments across various benchmarks including ImageNet~\cite{imagenet}, which are typical and widely used in the community.
We provide novel insights about both the strengths and weaknesses of widely adopted fine-tuning strategies for VLMs, from the perspective of the rationality of VLM predictions.
Moreover, our findings remain consistent across evaluation scenarios involving both in-distribution data and out-of-distribution data, as well as under various experimental settings.
This paper provides new insights for people to rethink the effects of mainstream fine-tuning methods for VLMs.

%% file: sec_ours/2_3_4.1_experimental_setup.tex
\section{Preliminaries}


There has been a surge of people exploring VLMs for their downstream tasks.
A typical way is to use them for image classification~\cite{flyp}.
In our prediction evaluations, we study the image classification task and measure model performances using the top-1 accuracy metric.

We evaluate whether the model provides valid evidence for its predictions by examining whether the explanation heatmap generated by VLMs focuses on the target objects. 
Specifically, a heatmap that strongly highlights key object regions while showing minimal responsiveness to background pixels indicates valid evidence.
Therefore, we rely on the ``Relevant Mass Accuracy (RMA)'' score~\cite{es_eval1, es_eval2}, which satisfies this criterion by measuring how much ``mass'' one method assigns to pixels within the region of target objects (ground truth).
RMA score is calculated by determining the ratio of the total heatmap pixel values within the target object regions, to the sum of all pixel values across the entire heatmap.
It requires both the generated explanation heatmap $(H)$ from VLMs and the ground truth explanation mask $(M)$, whose pixels on the target objects are marked as 1 otherwise marked as 0.
RMA score is defined as:
\begin{equation}
    \text{RMA}(H, M) = \frac{\sum H \odot M}{\sum H},
\label{es_eval}
\end{equation}
where $\odot$ represents Hadamard product.
Note that the evaluations from many studies~\cite{grad-cam, es_eval1} require the presence of ground-truth mask for heatmap localization.

\begin{figure}[t]
  \centering
  \includegraphics[width=0.46\textwidth]{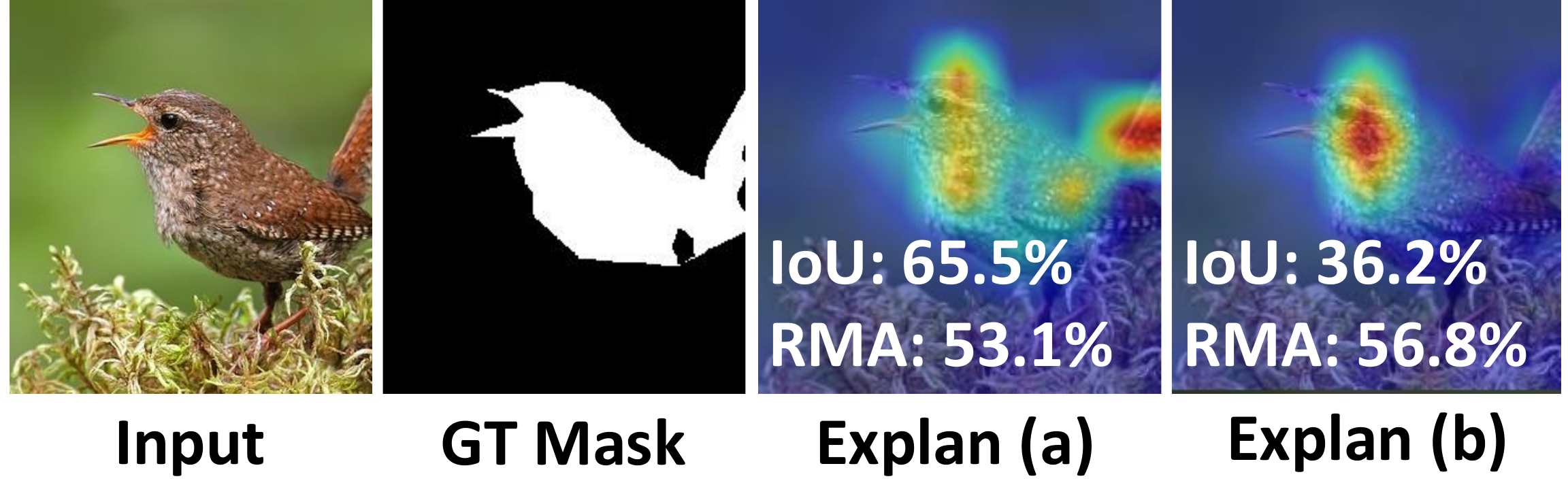}
  \caption{Both (a) and (b) have low responses to the background while (a) pays more attention to the whole body of the bird and (b) pays more attention to the discriminative feature of the bird (head). Compared with the IoU score between (a) and (b), the difference between them is negligible. Moreover, both achieve correct predictions. Input is from CUB-200-2011~\cite{cub} dataset. ``GT'' denotes abbreviation of ``Ground Truth'' and  ``Explan'' denotes abbreviation of ``Explanation''.}
  \label{rma_reliability}
\end{figure}


We emphasize that the RMA metric provides a more reasonable evaluation for classification tasks compared to metrics like ``Intersection over Union (IoU)" used in other works. 
For instance, Grad-CAM~\cite{grad-cam} relies on the IoU score to measure the overlap between the explanation heatmap and the ground truth mask. 
However, the IoU score fails to reasonably evaluate two vastly different yet valid pieces of evidence.
In Figure~\ref{rma_reliability}, we show two explanation heatmaps, (a) and (b), that are from different models.
Even though the IoU metric treats them differently, both of them achieve correct predictions with valid evidence.
They both exhibit a low response to background pixels.
(a) pays attention to the whole body of the bird.
(b) is also reasonable because it effectively identifies the distinguishing features of the bird, despite not highlighting the more complete bird region as in (a). 
This indicates that compared to IoU, RMA evaluation can fairly treat two distinct but valid evidence.

\textbf{Explanation Heatmap Generation.} 
The method we use is directly from ``Generic Attention Attribution''~\cite{chefer2021transformer_mm}.
In this case, the heatmaps are generated from attention maps of the transformer-based model, which is one of the most well-adopted methods, used in recent works including~\cite{rationale_2023_CVPR}. 
It has been demonstrated in existing work~\cite{liu2022rethinking} that it achieves the best faithfulness performance among all well-known explanation methods when applied to transformer-based models.
The main idea is Hadamard's product between attention maps and their gradient to the output.
It is defined as:
\begin{equation}
    \mathbf{\overline{A}} = \mathbb{E}_{h}((\mathbf{\triangledown A} \odot \mathbf{A})^+),
\label{heatmap_formula}
\end{equation}
where $\odot$ is the Hadamard product, $\mathbf{\triangledown A} := \frac{\partial y_t}{\partial \boldsymbol{A}}$ for $y_t$ which is the model’s output for the class $t$ that we wish to visualize.
$\mathbb{E}_{h}$ is the mean across the heads dimension.
The $+$ indicates that the negative contributions are removed before averaging.
Note that the class we explain are based on the index given by the annotations instead of predictions.



\section{Our Proposed Evaluations} 
We present our evaluation protocols with two criteria in mind.
(1) A trustworthy VLM should not produce instances of invalid evidence among samples with correct predictions.
(2) When focusing on the correct predicted objects, a reliable VLM should leverage such valid evidence to achieve correct predictions.
To determine whether the evidence (or rationale) of the model is correct, we use a threshold of 0.5 on the RMA measure. 
Specifically, an RMA score of 0.5 or above is considered valid evidence and vice versa.
As a result, we achieve four scenarios: RR, RW, WR, and WW (Figure~\ref{fig:model-safety}) that are used to formalize our two novel metrics:

\begin{figure}
    \centering
    \includegraphics[width=1.0\linewidth]{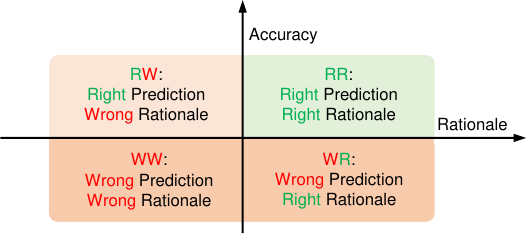}
    \caption{Overview of the four quadrants (RR, RW, WR, WW) of Accuracy and Rationale that are utilized to evaluate prediction rationality.}
    \label{fig:model-safety}
\end{figure}


1. \textbf{Prediction Trustworthiness (PT)}. 
A dependable and trustworthy model should generate valid evidence that corresponds to accurate predictions. 
Hence, we introduce the ``PT'' metric, which calculates the proportion of samples where the prediction is ``right'' and its evidence is also valid or ``right'' (RR) among all samples with right predictions, defined as:
\begin{equation}
    \text{PT} = \frac{{\text{RR}}}{{\text{RR} + \text{RW}}},
\label{formula_pt}
\end{equation}
where ``RW'' denotes data with the ``right'' classifications based on invalid or ``wrong'' evidence.
It is evident that an increase in the number of RW samples, i.e. irrational predictions, results in a decrease in PT scores.

2. \textbf{Inference Reliability (IR)}. 
Given that the model could pinpoint the regions of target objects, a reliable model should make correct predictions.
Consequently, we introduce the ``IR'' metric, which calculates the proportion of samples with correct prediction and valid evidence among all samples with valid evidence of target objects, defined as:
\begin{equation}
    \text{IR} = \frac{{\text{RR}}}{{\text{RR} + \text{WR}}},
\label{formula_ir}
\end{equation}
where ``WR'' denotes data with incorrect classifications with valid evidence.
An increase in the number of WR samples results in a decrease in IR scores.



\section{Experiments}

\subsection{Experimental Setup}
\textbf{Fine-tuning Methods.}
In this paper, we study fundamental methods including: (1) Zero-Shot (ZS), (2) Linear-Probing (LP), (3) Finetune Like CLIP Pretrain (FLCP), and (4) Fine-tuning (FT). 
For detailed information on these methods, please refer to our supplementary material in the extended version of our paper.


\textbf{Models.} 
We study four VLMs: the first two models are CLIP-ViT-B/32 \& 16~\cite{clip} from OpenAI, which manifest powerful zero-shot performances on image classification. 
The next two models are ALBEF-ViT-B/16~\cite{albef}, pretrained on 14M image-text pairs, and BLIP-ViT-B/16~\cite{blip}, pretrained on 129M image-text pairs, both developed by Salesforce. Their performances on the image classification task are also investigated in many works~\cite{satin, omnivl}.

\textbf{Fine-tuning Setups.} 
We maintain a consistent batch size and training epoch across all three fine-tuning methods (LP, FLCP, FT) for the same dataset and model. 
We employ the Adam~\cite{adam} optimizer during the fine-tuning. 
For more details about fine-tuning, please consult our supplementary material.

\textbf{Datasets.} 
In this paper, we conduct experiments on several datasets, including ImageNet~\cite{imagenet}, CalTech-101~\cite{caltect101}, Stanford-Dogs~\cite{standog}, CUB-200-2011~\cite{cub}, and ImageNet-C~\cite{imagenet-c}.
In CUB-200-2011 and CalTech-101 datasets, the 0-1 segmentation mask annotations directly serve as ground truth explanation masks.
For images with bounding box annotations surrounding predicted instances (ImageNet, ImageNet-C, Stanford-Dogs), we generate ground truth explanation masks as follows: given initial masks whose pixel values are all zero, we mark the mask areas within boxes as one.
For more detailed information about these datasets, please refer to the supplementary material.

%% file: sec_ours/4.2_weakness.tex
\subsection{Weaknesses of Fine-tuning}
\label{chap_weakness}

\input{tabs/tab_pred}

\input{tabs/tab_pt_ir_2}

\textit{Question: Will mainstream fine-tuning methods hurt the rationality of VLM predictions?}

\textit{Answer: Surprisingly yes! The well-adopted fine-tuning methods decrease the trustworthiness of VLM predictions in most settings: causing more samples with correct predictions based on invalid evidence.}

Although fine-tuning is able to improve the prediction accuracies of VLMs (see Table~\ref{tab_pred}), we find mainstream fine-tuning methods lead to worse prediction trustworthiness, as shown in Table~\ref{tab_ptir}.
For instance, in the ImageNet-1K dataset, with CLIP-ViT-B/16 model, compared with ZS, fine-tuning deteriorates ``Prediction Trustworthiness (PT)'' performances by $6.4\%$, $5.65\%$ and $4.07\%$ respectively.
Our experimental results confirm the significant drawbacks of mainstream fine-tuning methods for VLMs: fine-tuning results in more instances where predictions are correct but the evidence which VLMs base on is invalid.
This results in a reduced level of trustworthiness to VLM predictions.
Lastly, there are rare exceptions with increased PT scores. 
This is likely due to the low zero-shot prediction accuracy of ALBEF (12.43\%) and BLIP (16.88\%) on CUB-200-2011. 
Fine-tuning introduces the missing knowledge to these models, leading to increased PT.

\begin{figure*}[h]
  \centering
  \includegraphics[width=1.0\textwidth]{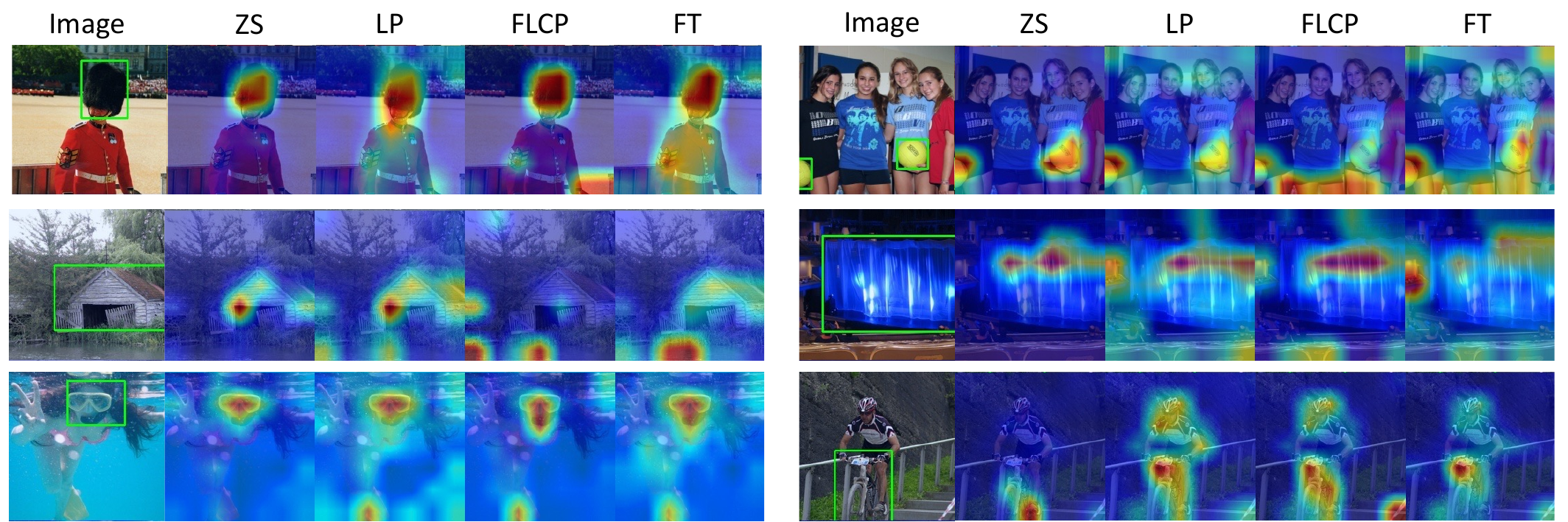}
  \caption{Visualization comparisons among different methods. Compared with zero-shot (ZS), current mainstream fine-tuning methods (LP, FLCP, and FT) for VLMs tend to show enhanced responses in background pixels that are irrelevant to predictions. Here we select the samples for which all four methods make correct predictions. Here we display bounding box annotations indicating the positions of the predicted target.}
  \label{fig_vis_id}
\end{figure*}

To further support our observation, we provide visualizations of the explanation heatmaps in Figure~\ref{fig_vis_id}.
We observe that widely adopted fine-tuning methods often amplify the responses of VLMs to pixels containing information irrelevant to the predicted objects.
For instance, from the leftmost first-row comparisons, fine-tuning makes VLMs enhance responses on the human body or background instead of the hat (predicted category).
Here we only show results on the CLIP-ViT-B/32 model with ImageNet-1K datasets due to space constraints.
Please refer to our supplementary material for more visualizations.

\textbf{Why does finetuning decrease trustworthiness?}
(1) VLMs tend to exploit the easiest path to minimize loss during finetuning, often picking up on spurious correlations or shortcuts present in the data. 
For instance, if all images of a particular class contain a common watermark or background, VLMs may associate that feature with the class label instead of learning the actual characteristics of the object. 
(2) Standard fine-tuning objectives usually prioritize improving prediction accuracy, but they do not account for the validity of the evidence used. 
As a result, there is no built-in mechanism to guide the model to focus on valid evidence.

In recent years, there have been some discussions regarding the excellence of fine-tuning for VLMs.
For example, existing work~\cite{flyp} shows that FLCP leads to uniformly better prediction performances.
They claim that FLCP should be adopted as the ``standard'' method for fine-tuning CLIP.
However, based on our discoveries, we contend that this conclusion doesn't apply when considering the rationality of VLM predictions. 
Although FLCP significantly enhances VLMs' prediction accuracies, we find that FLCP leads VLMs to provide more invalid evidence when making correct predictions, weakening the prediction trustworthiness of VLMs than ZS.
This disparity highlights the significance of considering different possibilities when evaluating VLMs' prediction rationality.

%% file: tabs/tab_pred.tex
\begin{table}[t]
\centering
\scriptsize
\begin{tabular}{ccccccc}
\hline
\multirow{2}{*}{Methods} & \multirow{2}{*}{VLMs} & \multicolumn{4}{c}{Datasets}                             & \multirow{2}{*}{Avg.}           \\ \cline{3-6}
                         &                       & IN & CT & SD & CUB &                                 \\ \hline
\multirow{4}{*}{ZS}      & ALBEF-ViT-B/16        & 46.48       & 77.02       & 29.25         & 12.43        & \multirow{4}{*}{53.74}          \\
                         & BLIP-ViT-B/16         & 46.30       & 85.89       & 32.38         & 16.88        &                                 \\
                         & CLIP-ViT-B/16         & 63.30       & 84.22       & 60.61         & 54.94        &                                 \\
                         & CLIP-ViT-B/32         & 58.41       & 84.79       & 54.62         & 52.33        &                                 \\ \hline
\multirow{4}{*}{LP}      & ALBEF-ViT-B/16        & 72.03       & 90.38       & 65.10         & 48.46        & \multirow{4}{*}{72.50}          \\
                         & BLIP-ViT-B/16         & 72.46       & 90.26       & 64.23         & 47.77        &                                 \\
                         & CLIP-ViT-B/16         & 76.69       & 94.64       & 74.14         & 70.14        &                                 \\
                         & CLIP-ViT-B/32         & 72.21       & 93.09       & 67.27         & 61.08        &                                 \\ \hline
\multirow{4}{*}{FLCP}    & ALBEF-ViT-B/16        & 77.58       & 95.85       & 77.88         & 77.27        & \multirow{4}{*}{\underline{80.99}}          \\
                         & BLIP-ViT-B/16         & 78.67       & 94.99       & 77.89         & 68.85        &                                 \\
                         & CLIP-ViT-B/16         & 72.41       & 96.20       & 80.70         & 80.76        &                                 \\
                         & CLIP-ViT-B/32         & 70.81       & 95.74       & 75.70         & 74.49        &                                 \\ \hline
\multirow{4}{*}{FT}      & ALBEF-ViT-B/16        & 80.82       & 95.91       & 81.32         & 80.48        & \multirow{4}{*}{\textbf{81.63}} \\
                         & BLIP-ViT-B/16         & 80.75       & 92.74       & 78.68         & 68.98        &                                 \\
                         & CLIP-ViT-B/16         & 81.19       & 93.03       & 81.56         & 79.25        &                                 \\
                         & CLIP-ViT-B/32         & 76.62       & 94.30       & 72.42         & 68.07        &                         \\ \hline       
\end{tabular}
\caption{
Comparisons of four methods regarding prediction accuracy (\%). 
The best-averaged score among the four methods is \textbf{bolded}, while the second-place averaged score is \underline{underlined}. 
Due to the space limit, we abbreviate the names of datasets.
Here, ``IN'', ``CT'', ``SD'', ``CUB'' denote ``ImageNet-1K'', ``CalTech-101'', ``Stanford-Dogs'', ``CUB-200-2011'' respectively.}
\label{tab_pred}
\end{table}

%% file: tabs/tab_pt_ir_2.tex
\begin{table*}[h]
\centering
\small
\begin{tabular}{cccccccc}
\toprule[1.2pt]
\multirow{2}{*}{Evaluations}                                                                  & \multirow{2}{*}{Methods} & \multirow{2}{*}{VLMs} & \multicolumn{4}{c}{Datasets}                             & \multirow{2}{*}{Avg.}  \\ \cline{4-7}
                                                                                              &                          &                       & ImageNet-1K & CalTech-101 & Stanford-Dogs & CUB-200-2011 &                        \\ \toprule[1.0pt]
\multirow{16}{*}{\begin{tabular}[c]{@{}c@{}}Prediction\\ Trustworthiness\\ (PT, \%) ↑\end{tabular}} & \multirow{4}{*}{ZS}      & ALBEF-ViT-B/16        & 90.61       & 76.28       & 95.02         & 49.31        & \multirow{4}{*}{\textbf{71.26}} \\
                                                                                              &                          & BLIP-ViT-B/16         & 89.01       & 61.72       & 93.95         & 23.93        &                        \\
                                                                                              &                          & CLIP-ViT-B/16         & 87.05       & 62.99       & 92.96         & 29.38        &                        \\
                                                                                              &                          & CLIP-ViT-B/32         & 89.39       & 73.44       & 94.58         & 30.57        &                        \\ \cline{2-8} 
                                                                                              & \multirow{4}{*}{LP}      & ALBEF-ViT-B/16        & 82.37       & 59.08       & 90.30         & 19.91        & \multirow{4}{*}{64.78} \\
                                                                                              &                          & BLIP-ViT-B/16         & 80.36       & 52.57       & 92.63         & 12.98        &                        \\
                                                                                              &                          & CLIP-ViT-B/16         & 80.65       & 56.40       & 92.19         & 36.17        &                        \\
                                                                                              &                          & CLIP-ViT-B/32         & 84.05       & 68.22       & 92.76         & 35.89        &                        \\ \cline{2-8} 
                                                                                              & \multirow{4}{*}{FLCP}    & ALBEF-ViT-B/16        & 87.07       & 62.43       & 92.68         & 64.33        & \multirow{4}{*}{\underline{67.95}} \\
                                                                                              &                          & BLIP-ViT-B/16         & 82.57       & 59.52       & 91.46         & 36.17        &                        \\
                                                                                              &                          & CLIP-ViT-B/16         & 81.40       & 64.32       & 76.44         & 16.56        &                        \\
                                                                                              &                          & CLIP-ViT-B/32         & 85.48       & 71.29       & 91.59         & 23.84        &                        \\ \cline{2-8} 
                                                                                              & \multirow{4}{*}{FT}      & ALBEF-ViT-B/16        & 86.28       & 48.97       & 92.22         & 24.98        & \multirow{4}{*}{67.01} \\
                                                                                              &                          & BLIP-ViT-B/16         & 85.54       & 39.96       & 93.13         & 25.85        &                        \\
                                                                                              &                          & CLIP-ViT-B/16         & 82.98       & 56.86       & 91.60         & 27.98        &                        \\
                                                                                              &                          & CLIP-ViT-B/32         & 86.29       & 80.01       & 94.17         & 55.43        &                        \\ \toprule[0.75pt]
\multirow{16}{*}{\begin{tabular}[c]{@{}c@{}}Inference\\ Reliability\\ (IR, \%) ↑\end{tabular}}      & \multirow{4}{*}{ZS}      & ALBEF-ViT-B/16        & 48.95       & 76.74       & 30.56         & 16.43        & \multirow{4}{*}{56.65} \\
                                                                                              &                          & BLIP-ViT-B/16         & 49.65       & 90.05       & 33.87         & 18.92        &                        \\
                                                                                              &                          & CLIP-ViT-B/16         & 66.33       & 85.58       & 61.96         & 68.05        &                        \\
                                                                                              &                          & CLIP-ViT-B/32         & 61.09       & 85.23       & 56.12         & 56.80        &                        \\ \cline{2-8} 
                                                                                              & \multirow{4}{*}{LP}      & ALBEF-ViT-B/16        & 74.76       & 92.56       & 66.21         & 55.46        & \multirow{4}{*}{75.67} \\
                                                                                              &                          & BLIP-ViT-B/16         & 74.78       & 90.89       & 65.11         & 59.91        &                        \\
                                                                                              &                          & CLIP-ViT-B/16         & 78.93       & 95.05       & 75.41         & 77.08        &                        \\
                                                                                              &                          & CLIP-ViT-B/32         & 74.91       & 93.76       & 68.53         & 67.37        &                        \\ \cline{2-8} 
                                                                                              & \multirow{4}{*}{FLCP}    & ALBEF-ViT-B/16        & 78.54       & 96.81       & 78.36         & 80.92        & \multirow{4}{*}{\underline{81.71}} \\
                                                                                              &                          & BLIP-ViT-B/16         & 80.04       & 94.93       & 78.73         & 73.40        &                        \\
                                                                                              &                          & CLIP-ViT-B/16         & 75.00       & 94.84       & 80.21         & 77.97        &                        \\
                                                                                              &                          & CLIP-ViT-B/32         & 71.84       & 95.46       & 76.43         & 73.87        &                        \\ \cline{2-8} 
                                                                                              & \multirow{4}{*}{FT}      & ALBEF-ViT-B/16        & 82.95       & 94.41       & 81.93         & 81.87        & \multirow{4}{*}{\textbf{83.52}} \\
                                                                                              &                          & BLIP-ViT-B/16         & 82.86       & 91.55       & 79.18         & 85.26        &                        \\
                                                                                              &                          & CLIP-ViT-B/16         & 83.25       & 90.66       & 82.06         & 81.12        &                        \\
                                                                                              &                          & CLIP-ViT-B/32         & 79.34       & 93.86       & 73.22         & 72.72        &                        \\ \bottomrule[1.2pt]
\end{tabular}
\caption{Comparisons of four methods with proposed ``PT'' and ``IR'' metrics. Here we observe that mainstream fine-tuning methods come with both strengths and weaknesses. We show that fine-tuning mostly leads to a worse capability of prediction trustworthiness but enhances the inference reliability of VLMs than the ZS method. The best-averaged score among the four methods is \textbf{bolded}, while the second-place averaged score is \underline{underlined}.}
\label{tab_ptir}
\end{table*}

%% file: sec_ours/4.3_strength.tex
\subsection{Strengths of Fine-tuning}
\label{chap_strength}

\textit{Question: Will valid evidence help enhance predictions made by fine-tuned VLMs?}


\textit{Answer: Yes, they exhibit good inference reliability; i.e., when focusing on the valid evidence of target objects, fine-tuned VLMs are more likely to make correct predictions.}

This phenomenon indicates better inference reliability of fine-tuning compared with ZS, as shown in Table~\ref{tab_ptir}.
For example, in the ImageNet-1K dataset, with the CLIP-ViT-B/16 model, LP, FLCP, and FT outperform ZS by $12.6\%$, $8.67\%$, and $16.92\%$ respectively; with the CLIP-ViT-B/32 model, LP, FLCP, and FT outperform ZS by $13.82\%$, $10.75\%$, and $18.25\%$ respectively.
This indicates that fine-tuning approaches contribute to less WR than ZS.
When VLMs identify valid evidence for target objects, fine-tuning is more likely to produce correct predictions.

Existing works~\cite{kumar2022calibrated, wortsman2022robust, flyp} are limited to the impact of mainstream VLM fine-tuning methods regarding predictive accuracies, ignoring their positive impacts on VLM prediction rationality. 
In this paper, we have analyzed and explored the benefits of fine-tuning VLMs from a new perspective.
Our experimental results show that fine-tuning has its merits and is not completely worthless for the prediction rationality of VLMs.

In summary, we conducted extensive experiments to validate the existing mainstream VLM fine-tuning methods in terms of both their strengths and weaknesses from a prediction rationality perspective. 
On the one hand, fine-tuning leads to good inference reliability: when provided with valid evidence of target objects, fine-tuned VLMs are more likely to generate accurate predictions.
On the other hand, we also confirm that mainstream fine-tuning methods tend to hurt the inherent capabilities of VLMs, specifically in terms of prediction trustworthiness.
These are aspects that merit attention from the community of machine learning.

%% file: sec_ours/4.4_ood.tex
\subsection{Analysis on Out-of-Distribution Data}
\label{chap_ood}

\textit{Question: Will out-of-distribution data change our observations?}


\textit{Answer: No, all findings remain consistent.}

Distributional shifts has garnered significant attention in the field of machine learning~\cite{fc_cvpr, fc_iclr}.
During the fine-tuning, the distributional discrepancy between the fine-tuning and testing data is worth considering.
Real-world data distributions can change due to factors such as time, location, and environment. 
Testing on out-of-distribution data helps simulate these changes, ensuring the model performs well in diverse scenarios. 
For example, in autonomous driving, the models need to remain stable in multiple weather conditions.

\begin{figure}[h]
  \centering
  \includegraphics[width=0.48\textwidth]{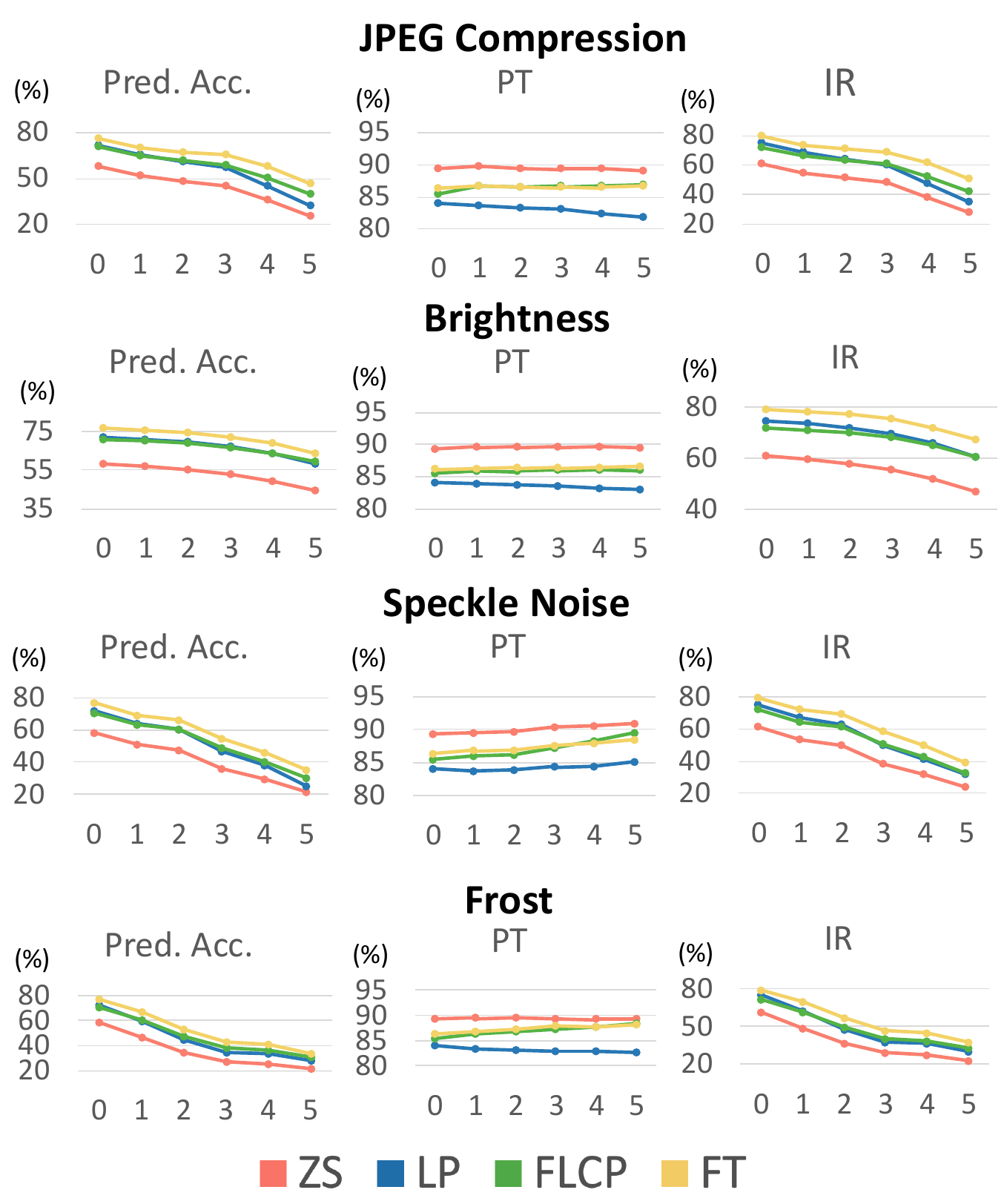}
  \caption{Experimental results on out-of-distribution data. Our discoveries remain consistent across various types and magnitudes of distributional shifts. The x-axis in all figures represents the strength of corruption, where a strength of 0 indicates the results of different methods on the original ImageNet validation data. Due to space constraints, we only show results with CLIP-ViT-B/32 and four types of corruption in the main paper. For more results, please refer to our supplementary material.}
  \label{fig_ood}
\end{figure}

In this section, we study the fine-tuning methods when testing on out-of-distribution data. 
Here we use the ImageNet-C dataset, which includes multiple corruption categories and levels of severity. 
As shown in Figure~\ref{fig_ood}, our key findings are as follows:
\begin{enumerate}
    \item Fine-tuning on in-distribution data can enhance the prediction accuracy for out-of-distribution data.
    \item However, the mainstream fine-tuning methods still compromise the prediction trustworthiness of VLM, which brings more samples with correct prediction based on invalid evidence, compared with zero-shot.
    \item Fine-tuning tends to enhance the inference reliability of VLMs: when focusing on correct prediction objects, fine-tuned VLMs are more likely to give correct predictions.
\end{enumerate}
Therefore, we extend our previous findings to scenarios involving out-of-distribution data, demonstrating the consistency of our discoveries.

Our conclusions also remain unaffected when the prediction accuracies decrease caused by corruption strength increases. 
Therefore, we think our findings may not change with variations in model prediction accuracy.

%% file: sec_ours/4.5_ablation_study.tex
\subsection{Ablations studies}
To ensure the consistency of our findings across different experimental settings, we perform a comprehensive series of ablation studies.
We investigate the effects under different setups including:
(1) Experiments with another popular optimizer: AdamW~\cite{adamw}.
(2) Experiments with another widely-used explanation method: gradient of attention ($\mathbf{\nabla A}$) based~\cite{grad_att} method.
The main idea of this method is to utilize the gradient of attention to the output as an explanation heatmap, where $\mathbf{\triangledown A} := \frac{\partial y_t}{\partial \boldsymbol{A}}$ for $y_t$ which is the model’s output for the class $t$.
(3) Results with different fine-tuning learning rates (abbreviated as ``LR''): $5e-4$ for ``LP'', $3e-6$ for ``FLCP'', and $2e-5$ for ``FT'', compared with the original setup, where we set learning rates as $1e-3$ for ``LP'', $5e-6$ for ``FLCP'', and $1e-5$ for ``FT''.
For the original learning rate settings regarding other models and datasets please refer to our supplementary material.
Note that the aforementioned three experiments are conducted with the CLIP-ViT-B/32 model on the ImageNet-1K.


As shown in Table~\ref{tab_ab}, \textit{our findings remain unaffected} with multiple setups. 
On the one hand, prevalent fine-tuning approaches tend to increase the instances with correct predictions based on invalid evidence, despite the enhancement in prediction accuracy. 
On the other hand, fine-tuning typically demonstrates strong inference reliability.

Recently, there have been other fine-tuning techniques proposed by the community including prompt tuning~\cite{coop}, and adapter tuning~\cite{tipa}.
We find that \textit{our findings are also consistent under these fine-tuning methods.} 
Due to the space limits please refer to our supplementary material for the related experimental results and introduction of these methods.

%% file: sec_ours/5_related_work.tex
\section{Related Works}
\label{sec:formatting}

\subsection{Multimodal Foundation Models}

In recent years, there has been a surge of interest in research regarding Vision-Language Models (VLMs). 
These VLMs~\cite{clip, albef, blip, flava, align, alignprompt, declip, florence, glip, otter, pali, regionclip, vilt, uniter}, have attracted substantial attention due to their remarkable capacity to achieve robust performance, both in zero-shot and fine-tuned scenarios, across a diverse spectrum of vision-language-related tasks~\cite{vqa, itr, ve, nlvr}.
Notably, CLIP~\cite{clip}, as a prominent exemplar in this domain, has also demonstrated exceptional zero-shot performance in image classification. 
The contrastive learning approach it employs has also found applications in fields such as multiview analysis~\cite{cmc} and egocentric video understanding~\cite{qitong_iccv}.
Recently, researchers have engaged in fine-tuning~\cite{flyp} VLMs to better adapt them to specific downstream tasks.
However, the impact of such training on the prediction rationality of these models remains an open research problem, one that warrants in-depth exploration and investigation.

\begin{table}[t]
\centering
\scriptsize
\begin{tabular}{c|c|cccc}
\hline
\multirow{2}{*}{Setup}                                                                              & \multirow{2}{*}{Evaluations} & \multicolumn{4}{c}{Methods}   \\ \cline{3-6} 
                                                                                                    &                              & ZS    & LP    & FLCP  & FT    \\ \hline
\multirow{3}{*}{\begin{tabular}[c]{@{}c@{}}AdamW \\ Optimizer\end{tabular}}                         & Pred. Acc.(\%) ↑             & 58.41 & 72.22 & 70.88 & \textbf{76.53} \\
                                                                                                    & PT(\%) ↑                     & \textbf{89.39} & 84.23 & 85.53 & 86.46 \\
                                                                                                    & IR(\%) ↑                     & 61.09 & 74.93 & 71.93 & \textbf{79.26} \\ \hline
\multirow{3}{*}{\begin{tabular}[c]{@{}c@{}}$\nabla A$ \\ Explanation \\ Heatmap\end{tabular}}       & Pred. Acc.(\%) ↑             & 58.41 & 72.21 & 70.81 & \textbf{76.62} \\
                                                                                                    & PT(\%) ↑                     & \textbf{74.79} & 63.62 & 65.21 & 65.76 \\
                                                                                                    & IR(\%) ↑                     & 61.18 & 75.18 & 71.87 & \textbf{79.68} \\ \hline
\multirow{3}{*}{\begin{tabular}[c]{@{}c@{}}Different LRs \\ Compared with \\ Original\end{tabular}} & Pred. Acc.(\%) ↑             & 58.41 & 72.28 & 70.05 & \textbf{75.51} \\
                                                                                                    & PT(\%) ↑                     & \textbf{89.39} & 84.72 & 86.19 & 86.59 \\
                                                                                                    & IR(\%) ↑                     & 61.09 & 74.91 & 71.21 & \textbf{78.62} \\ \hline
\end{tabular}
\caption{
Ablation studies with prediction accuracy, and our proposed ``Prediction Trustworthiness (PT)'' and ``Inference Reliability (IR)'' metrics.
Our findings are unaffected under different experimental setups.
The best score is bolded.}
\label{tab_ab}
\end{table}

\subsection{Explainable Machine Learning}

Explainable Machine Learning (XML) is crucial for promoting transparency, trust, accountability, and fairness in AI systems.
Researchers frequently employ techniques to explain neural network operations and decision-making regarding input data. 
Activation heatmaps such as Grad-CAM~\cite{grad-cam}, visualize important regions for specific classes. 
In light of the proliferation of transformer-based models~\cite{vit}, researchers start exploring the feasibility of utilizing attention maps, taking it as a way to provide explanations~\cite{chefer2021transformer_mm}.
In order to evaluate the quality of these explanation generation methods, existing works including~\cite{rise} study from the perspective of faithfulness; i.e., how accurately an explanation method reflects the true decision-making process of a model. 
In parallel, Mao et al.~\cite{rationale_2023_CVPR} propose the concept of a reliable model, emphasizing the importance of the "doubly-right" criterion: both accurate predictions and fine-grained language explanations of model decision-making.
Recently, some works~\cite{tang_nips, tang_eccv} have increasingly required VLM models to deliver not only accurate predictions but also correct rationales.
In this paper, we explore the impact of widely accepted fine-tuning methods on the prediction rationality of VLMs for vision tasks such as image classification, providing novel insights about VLM fine-tuning within the XML research community.
And we highlight that faithfulness is beyond the scope of our study due to two reasons.
On the one hand, faithfulness evaluations primarily focus on assessing the correctness of heatmap explanation methods.
On the other hand, existing work~\cite{liu2022rethinking} verified the superiority of our employed explanation generation method.

%% file: sec_ours/6_conclusion.tex
\section{Conclusion}

Prediction rationality is an important aspect to consider when fine-tuning Vision-Language Models (VLMs), especially in high-stakes applications. 
This paper provides a comprehensive assessment of the commonly used fine-tuning approaches, presenting some insights on both advantages and disadvantages.
On the one hand, they generally demonstrate strong inference reliability.
More specifically, when focusing on the valid evidence of target objects, the fine-tuned VLMs are more likely to make correct predictions.
On the other hand, fine-tuning often results in undermining the trustworthiness of VLM predictions by bringing more data samples with correct predictions based on invalid evidence.
We further observe that our discoveries are consistent across various types and magnitudes of distributional shifts, and remain unaffected with multiple setups.
To ensure that VLMs can be reliably used in high-stack applications, it will be crucial to study new fine-tuning methods that can improve VLM prediction rationality. 
We leave it as future works.
We expect our research may provide useful experience and advance the study of VLM fine-tuning.

\section{Acknowledgements}
This work is supported by the NSF CAREER Award No. 2340074, the NSF SLES Award No. 2416937, the NSF III CORE Award No. 2412675, and the DoD DEPSCoR Award AFOSR FA9550-23-1-0494. 
Any opinions, findings and conclusions or recommendations expressed in this material are those of the authors and do not reflect the views of the supporting entities.

%% file: sec_ours/suppl.tex

\clearpage

\section*{Supplementary Material}
This section contains supplementary material to support the main paper text.
It includes:
\begin{itemize}
\item Additional experimental results with out-of-distribution data. These include more corruption types and more VLMs compared with the main paper, which served as the extension of Figure~\ref{fig_ood}.
\item Visualizations with ALBEF-ViT-B/16, BLIP-ViT-B/16, and CLIP-ViT-B/16 models (Extension of Figure~\ref{fig_vis_id}).
\item More detailed description of the fine-tuning methods utilized in our main paper (Extension of paragraph titled ``Fine-tuning Methods'' in Section ``Experimental Setup'').
\item More detailed description of the datasets utilized in our study (Extension of paragraph titled ``Datasets'' in Section ``Experimental Setup'').
\item More implementation details of fine-tuning VLMs, which served as the extension of paragraph titled ``Fine-tuning Setups'' in Section ``Experimental Setup''.
\item Experiments with more fine-tuning techniques for VLMs, which served as the extension of Section ``Ablation Studies''.
\end{itemize}

\subsection*{Additional Results on Out-of-Distribution Data}

\begin{figure}[h]
  \centering
  \includegraphics[width=0.48\textwidth]{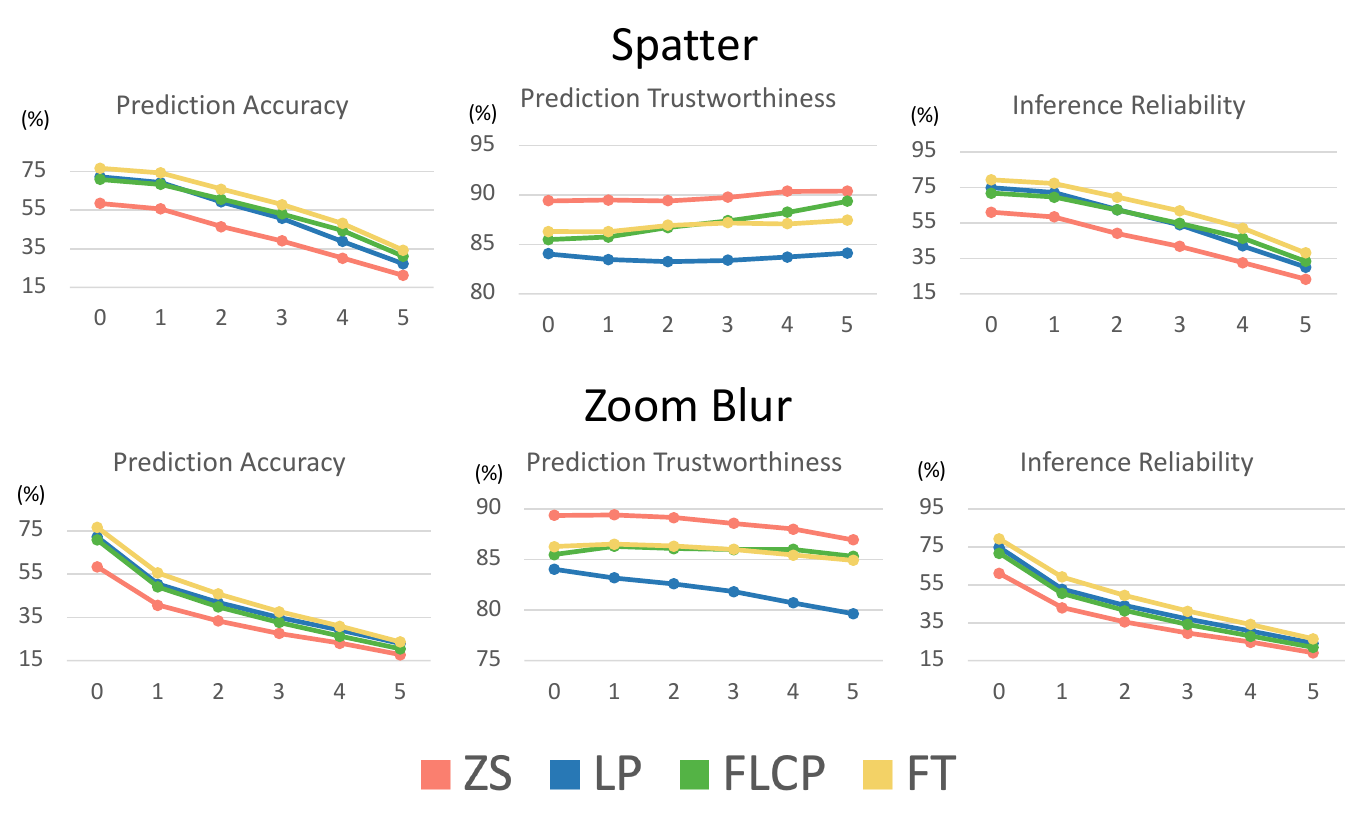}
  \caption{Additional results of testing on out-of-distribution data with CLIP-ViT-B/32 model.}
  \label{fig_ood_clip_b32_supp}
\end{figure}

\begin{figure*}[htbp]
  \centering
  \includegraphics[width=0.9\textwidth]{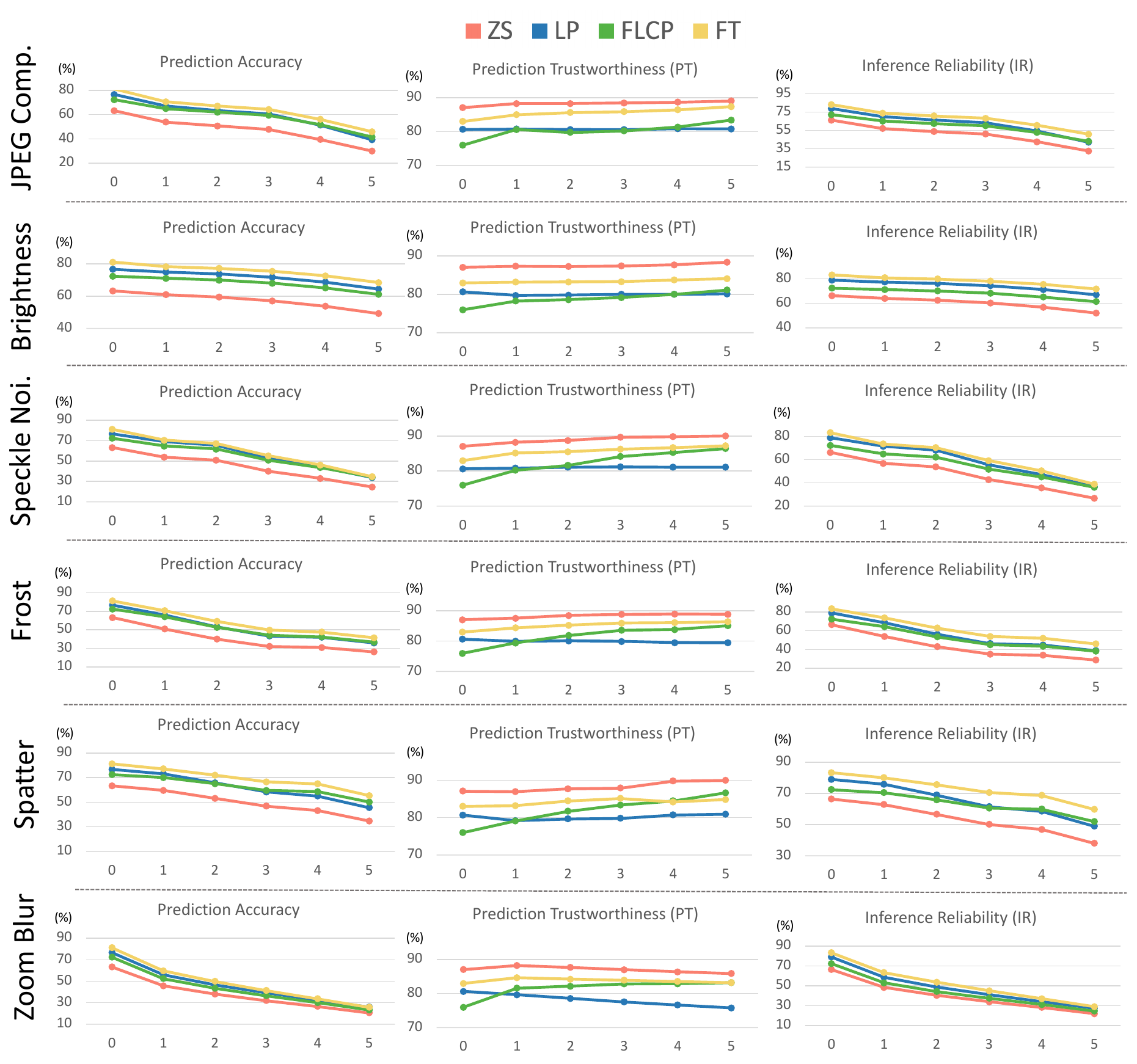}
  \caption{
  Experimental results on out-of-distribution data with CLIP-ViT-B/16 model.
  Due to the space limit in this figure, ``JPEG Compression'' and ``Speckle Noise'' are abbreviated as ``JPEG Comp.'' and ``Speckle Noi.'', respectively.}
  \label{fig_ood_clip_b16_supp}
\end{figure*}

Here, we utilize CLIP-ViT-B/16 and CLIP-ViT-B/32 models. 
For CLIP-ViT-B/16, we introduce six types of corruption from the ImageNet-C dataset.
Compared to the main paper's employment of CLIP-ViT-B/32 (see Figure~\ref{fig_ood}), we incorporate two additional types of corruption. 
As shown in Figure~\ref{fig_ood_clip_b32_supp} and Figure~\ref{fig_ood_clip_b16_supp}, with more corruption types and VLMs, our conclusions are consistent with those presented in the main paper.

\begin{figure*}[h]
  \centering
  \includegraphics[width=0.95\textwidth]{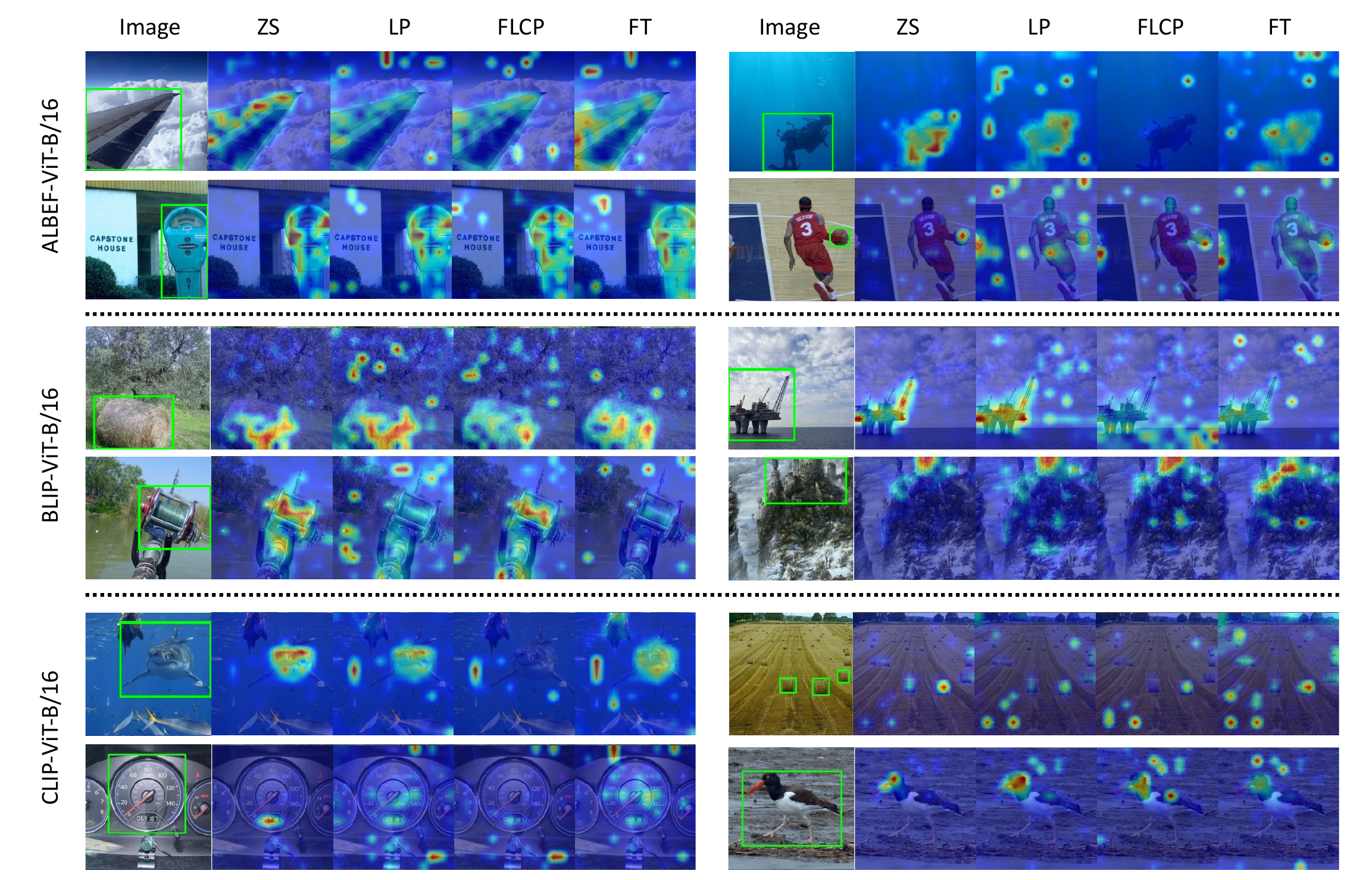}
  \caption{Additional visualizations among different methods on ImageNet-1K validation sets. Here we also select the samples for which all four methods make correct predictions. We display bounding box annotations indicating the target positions.}
  \label{fig_vis_id_supp}
\end{figure*}

\subsection*{Addditional Visualizations}

From Figure~\ref{fig_vis_id_supp}, we find that mainstream fine-tuning methods still deteriorate the prediction ratinoality of VLMs besides using the CLIP-ViT-B/32 model (shown in Figure~\ref{fig_vis_id}).
More specifically, fine-tuning tends to show enhanced responses to the background information, in contrast with ZS.
For example, from the rightmost second row, we find that fine-tuning methods show more responses to pixels (such as floor, and human) compared with ZS.
These pixels are irrelevant to the predicted category: ``basketball''.
Another example is that from the leftmost third row, we find that fine-tuning methods show more responses to pixels that are not related to the prediction object: ``hay''.
Our visualization evidence clarifies that compared with ZS VLMs, fine-tuned VLMs tend to base on unreasonable evidence even with correct predictions.

\input{tabs_supp/more_techs}

\subsection*{Detailed Information of Fine-tuning Methods}

1. \textbf{Zero-Shot (ZS)}.
The model is presented with images along with textual descriptions of target classes, where we follow~\cite{clip} and use the template: ``a photo of a $c_i$'' given k classes $\{c_1, c_2, ..., c_k\}$.
Here, we directly load the pretrained weights of the models and then evaluate by assigning the most similar class to each image among all classes in the dataset.

2. \textbf{Linear-Probing (LP)}.
In this context, we develop a neural network designed to process natural images as input data. 
This model comprises two distinct components.
The first component is the image encoder initialized with pre-trained weights obtained from the image encoder of VLMs. 
Importantly, the parameters of this image encoder are frozen throughout the fine-tuning process.
The second component is the classification head, responsible for making predictions, which is trainable during the fine-tuning phase.
During model training for each dataset, we utilize a loss function based on cross-entropy classification. 

3. \textbf{Finetune Like CLIP Pretrain (FLCP)}.
We load the pre-trained weights of VLMs. 
We align the images along with textual descriptions of target classes in a contrastive manner during fine-tuning for each dataset, following the same training scenarios as the pretaining process of CLIP~\cite{clip}.
And we follow the same evaluation protocol as ZS after fine-tuning.

4. \textbf{Fine-tuning (FT)}.
In this case, we build the same neural network as LP method, with the same initialization and evaluation scenarios.
The only difference is that the image encoder is trainable when fine-tuning the model.

\subsection*{Detailed Information of Datasets}

We use datasets that are widely used in the community. 
The specific information about these datasets is as follows:

\textbf{ImageNet}~\cite{imagenet} is a large-scale image database that has played an important role in computer vision and deep learning research.
Here we access \textbf{ImageNet-1K} which is one of the most commonly used subsets of ImageNet. 
\textbf{ImageNet-1K} is a large-scale image database that has played an important role in computer vision and deep learning research.
It spans 1000 object classes and contains 1,281,167 training images, 50,000 validation images, and 100,000 test images.
The bounding boxes for instances in this dataset are included.
We utilize the validation set during evaluations. 

\textbf{CalTech-101}~\cite{caltect101} contains pictures of objects belonging to 101 categories, which includes 40 to 800 images per category. 
The segmentation masks for instances in this dataset are included. 
The size of each image is roughly $300\times200$ pixels.
We partition the dataset into a hold-out test set, consisting of 20\% of the data, while the remaining 80\% will be used for fine-tuning VLMs.

\textbf{Stanford-Dogs}~\cite{standog} contains images of 120 types of dogs from around the world. 
This dataset has been built using images and annotation from ImageNet for the task of fine-grained image categorization.
It contains 20,580 images in total (12,000 for training and 8,580 for testing) and includes class labels, and bounding boxes for annotations.

\textbf{CUB-200-2011}~\cite{cub} is one of the most widely used datasets for fine-grained visual categorization tasks. 
It contains 11,788 images of 200 bird categories, 5,994 for training, and 5,794 for testing. 
Each image has one category label and one bird segmentation mask annotation.

\textbf{ImageNet-C}~\cite{imagenet-c} is an open-source dataset that consists of algorithmically generated corruptions (such as blur, noise) applied to the ImageNet validation set.
Similar to ImageNet-1K, each image is accompanied by a single bounding box annotation that delineates the instances present within images.
Unlike the four previously mentioned datasets, we only employ this dataset for evaluations of out-of-distribution data.

\subsection*{More VLM Finetuning Details}

For all three studied fine-tuning methods (LP, FLYP, FT) in main paper, besides ImageNet-1K, where we fine-tune with four A6000 Nvidia GPUs, we fine-tune VLMs using one A6000 Nvidia GPU for the rest of the datasets.
We set the batch size as 128 per GPU for the ImageNet-1K dataset, and 64 for the CalTech-101 and Stanford-Dogs datasets when finetuning.
For the CUB-200-2011 dataset, We set the batch size as 64 when fine-tuning CLIP models and 128 when fine-tuning ALBEF and BLIP models.
We set the training epochs as 10 for the ImageNet-1K and Stanford-Dogs datasets.
For the CalTech-101 dataset, We set the training epochs as 10 when fine-tuning CLIP models and 5 when fine-tuning ALBEF and 3 when fine-tuning BLIP models.
For the CUB-200-2011 dataset, We set the training epochs as 20 when fine-tuning the ALBEF model and 10 when fine-tuning the rest of the models.

\begin{table}[h]
\centering
\scriptsize
\begin{tabular}{cccccc}
\hline
\multirow{2}{*}{Methods} & \multirow{2}{*}{VLMs} & \multicolumn{4}{c}{Datasets}                             \\ \cline{3-6} 
                         &                       & IN & CT & SD & CUB \\ \hline
\multirow{4}{*}{FLCP}    & ALBEF-ViT-B/16        & 1e-5        & 3e-5        & 3e-5          & 3e-5         \\
                         & BLIP-ViT-B/16         & 1e-5        & 3e-5        & 3e-5          & 1e-5         \\
                         & CLIP-ViT-B/16         & 1e-6        & 3e-6        & 1e-5          & 1e-5         \\
                         & CLIP-ViT-B/32         & 5e-6        & 2e-5        & 1e-5          & 1e-5         \\ \hline
\multirow{4}{*}{FT}      & ALBEF-ViT-B/16        & 3e-5        & 1e-4        & 3e-5          & 3e-5         \\
                         & BLIP-ViT-B/16         & 3e-5        & 1e-4        & 3e-5          & 2e-5         \\
                         & CLIP-ViT-B/16         & 1e-5        & 3e-5        & 1e-5          & 1e-5         \\
                         & CLIP-ViT-B/32         & 1e-5        & 1e-5        & 1e-5          & 2e-5         \\ \hline
\end{tabular}
\caption{Learning rate settings when fine-tuning with FLCP and FT methods.}
\label{supp_lr}
\end{table}

For all experiments, we use Python-3.9.16, PyTorch-1.9.1, TorchVision-0.10.1, and cudatoolkit-11.4.2.
And the hyperparameters of the Adam optimizer are all set as follows: betas$=(0.9, 0.999)$, eps$=1e-8$, weight decay$=0.0$.
For the FLYP method, the temperature value of the contrastive loss is set to $0.07$.
The learning rate of the LP method is set as $1e-3$. The learning rate settings of FLYP and FT methods as shown in Table~\ref{supp_lr}.
The reason why learning rates behave in variations under different training scenarios is that we find the appropriate learning rate varies in different situations.
For example, when fine-tuning on ImageNet-1K, the learning rate with $1e-3$ is too big for FLYP and FT methods to let models converge during training.
We provide the results from a single run. We did not observe any variation in results with the same setup.

\subsection*{Experiments with More Fine-tune Techniques}

Recently, there have been other fine-tuning techniques proposed by the community.
Typical ones include prompt tuning such as CoOp~\cite{coop}, adapter tuning such as Tip-Adapter~\cite{tipa}.
Here we experiment with these three techniques (CoOp, Tip-Adapter) with CLIP-ViT-B/16 and CLIP-ViT-B/32 models.
The high-level concepts of these methods are introduced here: 

1. \textbf{CoOp} uses learnable vectors to model the words in the prompt, while keeping the parameters of the pre-trained VLM fixed throughout the process. 
It considers two types of learnable prompts: the first is a unified context, where the learnable context is the same regardless of the sample's category; the second is a class-specific context, where each category has its own unique learnable context.

2. \textbf{Tip-Adapter} leverages VLM such as CLIP to construct a cache model, which stores classification knowledge from downstream training data. 
Based on this approach, Tip-Adapter-F turns the Keys part of the Cache Model into learnable parameters, allowing them to be updated through training.
Here we employ the Tip-Adapter-F method in our study.


As shown in Table~\ref{tab_more_tech}, we find our findings remain consistent.
While these fine-tuning strategies (CoOp, Tip-Adapter) tend to enhance the inference reliability (IR) of VLMs, they often deteriorate the prediction trustworthiness (PT).


%% file: tabs_supp/more_techs.tex
\begin{table*}[h]
\centering
\begin{tabular}{cccccccc}
\toprule[1.2pt]
\multirow{2}{*}{Evaluations}                                                            & \multirow{2}{*}{Methods} & \multirow{2}{*}{VLMs} & \multicolumn{4}{c}{Datasets}                             & \multirow{2}{*}{Avg.}  \\ \cline{4-7}
                                                                                        &                          &                       & ImageNet-1K & CalTech-101 & Stanford-Dogs & CUB-200-2011 &                        \\ \bottomrule[1.0pt]
\multirow{6}{*}{\begin{tabular}[c]{@{}c@{}}Prediction \\ Accuracy (\%) ↑\end{tabular}}  & \multirow{2}{*}{ZS}      & CLIP-ViT-B/16         & 63.30       & 84.22       & 60.61         & 54.94        & \multirow{2}{*}{64.15}  \\
                                                                                        &                          & CLIP-ViT-B/32         & 58.41       & 84.79       & 54.62         & 52.33        &                        \\ \cline{2-8} 
                                                                                        & \multirow{2}{*}{CoOp}      & CLIP-ViT-B/16         & 76.06       & 92.17       & 73.54         & 68.17        & \multirow{2}{*}{75.80}  \\
                                                                                        &                          & CLIP-ViT-B/32         & 71.21       & 92.28       & 68.95         & 64.03        &                        \\ \cline{2-8} 
                                                                                        & \multirow{2}{*}{TA}      & CLIP-ViT-B/16         & 76.34       & 93.14       & 72.31         & 75.27       & \multirow{2}{*}{\textbf{76.73}}  \\ 
                                                                                        &                          & CLIP-ViT-B/32         & 71.94       & 93.09       & 67.18         & 64.57        &                        \\ \toprule[0.75pt] 
\multirow{6}{*}{\begin{tabular}[c]{@{}c@{}} Prediction\\ Trustworthiness\\ (PT, \%) ↑\end{tabular}}  & \multirow{2}{*}{ZS}      & CLIP-ViT-B/16         & 87.05       & 62.99       & 92.96         & 29.38       & \multirow{2}{*}{\textbf{70.05}}  \\
                                                                                        &                          & CLIP-ViT-B/32         & 89.39       & 73.44       & 94.58         & 30.57        &                        \\ \cline{2-8}
                                                                                        & \multirow{2}{*}{CoOp}      & CLIP-ViT-B/16         & 75.35       & 35.94       & 83.25         & 2.35        & \multirow{2}{*}{54.43}  \\
                                                                                        &                          & CLIP-ViT-B/32         & 79.59       & 47.96       & 90.30         & 20.70        &                        \\ \cline{2-8} 
                                                                                        & \multirow{2}{*}{TA}      & CLIP-ViT-B/16         & 66.76       & 37.90       & 85.95         & 6.14       & \multirow{2}{*}{52.24}  \\
                                                                                        &                          & CLIP-ViT-B/32         & 72.77       & 50.25       & 89.19         & 8.98        &                        \\ \toprule[0.75pt] 
\multirow{6}{*}{\begin{tabular}[c]{@{}c@{}} Inference\\ Reliability\\ (IR, \%) ↑\end{tabular}}  & \multirow{2}{*}{ZS}      & CLIP-ViT-B/16         & 66.33       & 85.58       & 61.96         & 68.05        & \multirow{2}{*}{67.65}  \\
                                                                                        &                          & CLIP-ViT-B/32         & 61.09       & 85.23       & 56.12         & 56.80        &                        \\ \cline{2-8}
                                                                                        & \multirow{2}{*}{CoOp}      & CLIP-ViT-B/16         & 78.50       & 96.13       & 75.41         & 69.93        & \multirow{2}{*}{78.84}  \\
                                                                                        &                          & CLIP-ViT-B/32         & 73.90       & 92.82       & 70.99         & 73.00        &                        \\ \cline{2-8} 
                                                                                        & \multirow{2}{*}{TA}      & CLIP-ViT-B/16         & 78.42       & 96.06       & 74.39         & 86.73       & \multirow{2}{*}{\textbf{81.00}}  \\
                                                                                        &                          & CLIP-ViT-B/32         & 74.60       & 94.50       & 69.09         & 74.17        &                        \\ \bottomrule[1.2pt]
\end{tabular}
\caption{
Comparisons of three methods regarding prediction accuracy and our proposed PT and IR scores.
Tip-Adapter is abbreviated as ``TA'' due to the space limit in this table.
The best averaged scores are marked as \textbf{bold}.
In this case, these fine-tuning methods (CoOp, Tip-Adapter) all tend to outperform ZS in prediction accuracy and IR.
However, ZS achieves the best averaged performances with PT evaluations.
}
\label{tab_more_tech}
\end{table*}